# Automatic Detection of Blue-White Veil and Related Structures in Dermoscopy Images

M. Emre Celebi[1,♣], Hitoshi Iyatomi[2], William V. Stoecker[3], Randy H. Moss[4], Harold S. Rabinovitz[5], Giuseppe Argenziano[6], H. Peter Soyer[7]

ecelebi@lsus.edu, iyatomi@hosei.ac.jp, wvs@mst.edu, rhm@mst.edu, harold@admcorp.com, argenziano@tin.it

peter.soyer@telederm.eu

[1] Department of Computer Science, Louisiana State University in Shreveport, Shreveport, LA, USA

[2] Department of Electrical Informatics, Hosei University, Tokyo, Japan

[3] Stoecker & Associates, Rolla, MO, USA

[4] Department of Electrical and Computer Engineering, Missouri University of Science and Technology, Rolla, MO, USA

[5] Department of Dermatology, University of Miami School of Medicine, Miami, Florida, USA.

[6] Department of Dermatology, Second University of Naples, Naples, Italy

[7] Queensland Institute of Dermatology, University of Queensland, Woolloongaba, Australia.

---

[♣] Contact author (Email: ecelebi@lsus.edu)

# ABSTRACT

Dermoscopy is a non-invasive skin imaging technique, which permits visualization of features of pigmented melanocytic neoplasms that are not discernable by examination with the naked eye. One of the most important features for the diagnosis of melanoma in dermoscopy images is the blue-white veil (irregular, structureless areas of confluent blue pigmentation with an overlying white "ground-glass" film). In this article, we present a machine learning approach to the detection of blue-white veil and related structures in dermoscopy images. The method involves contextual pixel classification using a decision tree classifier. The percentage of blue-white areas detected in a lesion combined with a simple shape descriptor yielded a sensitivity of 69.35% and a specificity of 89.97% on a set of 545 dermoscopy images. The sensitivity rises to 78.20% for detection of blue veil in those cases where it is a primary feature for melanoma recognition.

## 1. INTRODUCTION

Malignant melanoma, the most deadly form of skin cancer, is one of the most rapidly increasing cancers in the world, with an estimated incidence of 59,940 and an estimated total of 8,110 deaths in the United States in 2007 alone [1]. Dermoscopy is a non-invasive skin imaging technique which permits visualization of features of pigmented melanocytic neoplasms that are not discernable by examination with the naked eye. Practiced by experienced observers, this imaging modality offers higher diagnostic accuracy than observation without magnification [2-5]. Dermoscopy allows the identification of dozens of morphological features one of which is the blue-white veil (irregular, structureless areas of confluent blue pigmentation with an overlying white "ground-glass" film) [6]. This feature is one of the most significant dermoscopic indicator of invasive malignant melanoma, with a sensitivity of 51% and a specificity of 97% [7]. Figure 1 shows a melanoma with blue-white veil.

**INSERT FIGURE 1 HERE**

Numerous methods for extracting features from clinical skin lesion images have been proposed in the literature [8-10]. However, feature extraction in dermoscopy images is relatively unexplored. The dermoscopic feature extraction studies to date include two pilot studies on pigment networks [11][12] and globules [11], and three systematic studies on dots [13] and blotches [14][15]. To the best of our knowledge, there is no published systematic study on the detection of blue-white veil.

In this article, we present a machine learning approach to the detection of blue-white veil in dermoscopy images. Figure 2 shows an overview of the approach. The rest of the paper is organized as follows. Section 2 describes the image set and the preprocessing phase. Section 3 discusses the feature extraction. Section 4 presents the pixel classification. Section 5 describes the classification of lesions based on the blue-white veil feature. Finally, Section 6 gives the conclusions.

**INSERT FIGURE 2 HERE**

## 2. IMAGE SET DESCRIPTION AND PREPROCESSING

### 2.1. Image Set Description

The image set used in this study consists of 545 digital dermoscopy images obtained from two atlases. The first is the CD-ROM Interactive Atlas of Dermoscopy [6], which is a collection of images acquired in three institutions: University Federico II of Naples, Italy, University of Graz, Austria, and University of Florence, Italy. The second atlas is a pre-publication version of the American Academy of Dermatology DVD on Dermoscopy, edited by Harold Rabinovitz et al. These were true-color images with a typical resolution of 768 x 512 pixels. The diagnosis distribution of the cases was as follows: 299 dysplastic nevi, 186 melanomas, 28 blue nevi, 14 Reed/Spitz nevi, 8 combined nevi, 8 basal cell carcinoma, and 2 intradermal nevi. The lesions were biopsied and diagnosed histopathologically in cases where significant risk for melanoma was present; otherwise they were diagnosed by follow-up examination.

## 2.2. Preprocessing

Prior to the feature extraction two preprocessing steps, namely the determination of the background skin color and selection of the training and test pixels, were performed on the images. Figure 3 shows an overview of this procedure.

**INSERT FIGURE 3 HERE**

The lesion borders were obtained manually under the supervision of an experienced dermatologist (WVS). The motivation for using manual borders rather than computer-detected borders [16][17] was to separate the problem of feature extraction from the problem of automated border detection. The procedure for manual border determination was as follows. First, a number of points were selected along the lesion border. These points were then connected using a second-order B-spline function. Finally, the resulting closed curve was filled using a flood-fill algorithm to obtain the binary border mask. Figure 4 a-b illustrates this procedure.

For the extraction of the color features, the background skin color needs to be determined. First, the region outside the border with an area equal to 10% of the lesion area was omitted to reduce the effects of peripheral inflammation and errors in border determination. The background skin color was then calculated as the average color over the next region outside the border with an area equal to 20% of the lesion area. The non-skin pixels (black image frames, rulers, hairs, and bubbles) were not included in the calculation. The omitted pixels were those that are determined not to satisfy the following empirical rule [18]: ( $R > 90 \cap R > B \cap R > G$ ), where $R$, $G$, and $B$ denote the red, green, and blue values, respectively, of the pixel under consideration. The 10% and 20% areas outside the lesion were determined from the binary border mask using the Euclidean distance transform. Figure 4c shows these areas for a sample lesion.

In order to select training and test pixels for classification, 100 images were chosen from the entire image set. Forty-three of these images had sizeable pure veil regions and 62 had sizeable pure non-veil regions. In each image, a number of small circular regions that contain either veil or non-veil pixels were manually determined. Training and test pixels were then randomly selected from these manually determined regions. The selection method was designed to ensure a balanced distribution of the two classes (veil and non-veil) in the training set [19]. Figure 4d shows two manually selected regions on a sample image.

For each lesion, two additional features, primary blue-white veil and veil-related structures, were determined by a dermatologist (WVS). A feature such as a veil is said to be a primary feature if the veil is the feature most characteristic of melanoma, i.e. the feature present in the lesion which is most recognizable and specific for melanoma. Some structures related to blue-white veils were also considered in this study. These included gray or blue-gray veils or any veils which lacked the whitish film seen in the classic veil. These were identified as veil-related structures.

**INSERT FIGURE 4 HERE**

### 3. FEATURE EXTRACTION

After the selection of training and test pixels, features that will be used in the classification of these pixels need to be extracted. There are two main approaches to pixel classification: non-contextual and contextual [20]. In non-contextual pixel classification, during feature extraction, a pixel is treated in isolation from its spatial neighborhood. This often leads to noisy results. On the other hand, in contextual pixel classification, the spatial neighborhood of the pixel is also taken into account. In this study, the latter approach is followed. Several features were extracted in the 5 x 5 neighborhood of each pixel. For each feature, the median value in the neighborhood was then taken as the value for that feature of the center pixel. To speed up the median search in a 5 x 5 neighborhood, instead of fully sorting the 25 values, a minimum exchange network algorithm that performs a partial sort was employed [21]. Fifteen color features and three texture features were used to characterize the image pixels.

### 3.1. Absolute color features

The absolute color of a pixel was quantified by its chromaticity coordinates $F_1$, $F_2$, and $F_3$ (see Table 1). An advantage of $F_1$, $F_2$, and $F_3$ over the raw $R$, $G$, and $B$ values is that while the former are invariant to illumination direction and intensity [22], the latter are not. This invariance is essential for dealing with images that are acquired in uncontrolled imaging conditions.

### 3.2. Relative color features

Relative color refers to the color of a lesion pixel when compared to the average color of the background skin. A total of 12 relative color features were extracted from each pixel (see Table 1). In the table, the lesion pixel and the average background skin color in the *RGB* color space are denoted as ($R_L$, $G_L$, $B_L$) and ($R_S$, $G_S$, $B_S$), respectively. The relative color features offer several advantages. First, they compensate for variations in the images caused by illumination and/or digitization. Second, they equalize variations in normal skin color among individuals. Third, relative color is more natural from a perceptual point of view. Recent studies [15][23][24] have confirmed the usefulness of relative color features in skin lesion image analysis.

### 3.3. Texture features

In order to quantify the texture in the 5 x 5 neighborhood of a pixel, a set of statistical texture descriptors based on the Gray Level Co-occurrence Matrix (*GLCM*) were employed [25]. Although many statistics can be derived from the *GLCM*, three gray-level shift-invariant statistics (entropy $F_{16}$, contrast $F_{17}$, and correlation $F_{18}$) were used in this study to obtain a non-redundant texture characterization that is robust to linear shifts in the illumination intensity [26]. In order to achieve rotation invariance, the normalized *GLCM* was computed for each of the 4 directions $\{0°, 45°, 90°, 135°\}$ and the statistics calculated from these matrices were averaged.

**INSERT TABLE 1 HERE**

## 4. PIXEL CLASSIFICATION

Popular classifiers used in pixel classification tasks include k-nearest neighbor [27], Bayesian [28], artificial neural networks [29], and support vector machines [28]. In this study, a decision tree classifier was used to classify the image pixels into 2 classes: veil and non-veil. The motivation for this choice was two-fold. First, decision tree classifiers generate easy-to-understand rules, which is important for the clinical acceptance of a computer-aided diagnosis system. Second, they are fast to train and apply. The well-known *C4.5* algorithm [30] was used for decision tree induction.

Given a large training set, decision tree classifiers, in general, generate complex decision rules that perform well on the training data, but do not generalize well to unseen data [31]. In such cases, the classifier model is said to have overfit the training data. The *C4.5* algorithm prevents overfitting by pruning the initial tree that is, by identifying subtrees that contribute little to predictive accuracy and replacing each by a leaf [30]. The confidence factor (*C*) parameter controls the level of pruning and has a default value of 0.25. Another parameter that influences the complexity of the induced tree is the minimum number of samples per leaf (*M*). The default value for *M* is 2. In order to induce a simple tree that generalizes better, *C* and *M* were set to 0.1 and 100, respectively. Using these parameter values, the *C4.5* algorithm was trained with the manually selected training pixels (Section 2.2). Figure 5 shows the induced decision tree. It can be seen that only 2 of the 18 features were included in the classification model. One of these is an absolute color feature ($F_3$), whereas the other one is a relative color feature ($F_{10}$). The classification performance of the tree on the manually selected test pixels was a sensitivity (percentage of correctly detected veil pixels) of 84.33% and a specificity (percentage of correctly detected non-veil pixels) of 96.19%.

**INSERT FIGURE 5 HERE**

In order to evaluate the effectiveness of the classification model, the induced decision rules were applied to the entire image set. In the classifier training phase, 18 features were extracted from the training pixels. In contrast, in the rule application phase, only the two features that appear in the decision tree, namely $F_3$ and $F_{10}$, need to be extracted from the pixels. For each image, an initial binary veil mask was generated as a result of the rule application. To smooth the borders, a 5 x 5 majority filter [32] was applied to the initial masks. This filter replaces each pixel's value with the majority class label in its 5 x 5 neighborhood. Figure 6 shows the initial and final veil masks for a sample image.

**INSERT FIGURE 6 HERE**

Figure 7 shows a sample of the detection results. In this figure, parts (a) through (f) are melanomas, (g) is a Reed/Spitz nevus, and (h) is a blue nevus. It can be seen that the presented method detects most of the blue-white areas accurately.

**INSERT FIGURE 7 HERE**

**5. LESION CLASSIFICATION BASED ON THE BLUE-WHITE VEIL FEATURE**

In the second part of the study, we developed a second classifier to discriminate between melanoma and benign lesions based on the presence/absence of the blue-white veil feature. In order to characterize the detected blue-white areas, we used a numerical feature defined as follows:

$$S_1 = \frac{Area\ of\ Detected\ Blue\ White\ Veil}{Area\ of\ Lesion} \quad (1)$$

The problem with using $S_1$ alone is that a blue nevus (such as the one in Fig7h) might be misclassified as melanoma due to its high percentage of blue-white areas. We can solve this problem by using additional features that characterize the circularity and/or ellipticity of the lesion. The circularity of a lesion can be characterized by [33]:

$$S_2 = \frac{m_R}{\sigma_R} = \frac{\frac{1}{P}\sum_{k=1}^{P}\|(r_k,c_k)-(\overline{r},\overline{c})\|}{\left(\frac{1}{P}\sum_{k=1}^{P}(\|(r_k,c_k)-(\overline{r},\overline{c})\|-m_R)^2\right)^{1/2}} \quad (2)$$

where $P$ is the number of points on the lesion boundary, $(r_k, c_k)$ is the spatial coordinate of the $k^{th}$ boundary point, and $(\bar{r}, \bar{c})$ is the centroid of the lesion object (see Fig. 4b). The ellipticity of a lesion can be measured by [34]:

$$S_3 = \begin{cases} 16\pi^2 A_1 & \text{if } A_1 \leq \dfrac{1}{16\pi^2} \\ \dfrac{1}{16\pi^2 A_1} & \text{otherwise} \end{cases} \quad (3)$$

$$\mu_{pq} = \sum_{i=0}^{Nr} \sum_{j=0}^{Nc} I(i,j) \cdot (i-\bar{r})^p \cdot (j-\bar{c})^q$$

$$A_1 = \frac{\mu_{20}\mu_{02} - \mu_{11}^2}{\mu_{00}^4}$$

where, $I$ is the binary lesion image (see Fig. 4b), $Nr$ and $Nc$ are the number of rows and number of columns in $I$, respectively.

The rationale behind the inclusion of $S_2$ and $S_3$ is that benign lesions with blue-white areas might be distinguished from melanomas by their highly circular ($S_2$) and/or elliptical ($S_3$) shapes. As in the pixel classification procedure, we used the *C4.5* algorithm with 10-fold cross-validation to generate a classification model based on the features $S_1$, $S_2$, and $S_3$. Figure 8 shows the induced decision tree.

**INSERT FIGURE 8 HERE**

As expected, a lesion is classified as benign if it contains none to very small, i.e. less than 0.9%, blue-white areas. On the other hand, if the lesion contains significantly large blue-white areas, the ellipticity value is checked. If the lesion is highly elliptical, i.e. the $S_3$ value is greater than 0.979, then it is classified as benign; otherwise, it is classified as melanoma. Note that, the circularity feature ($S_2$) was discarded by the induction algorithm possibly because its characteristics are captured by the more general ellipticity feature ($S_3$).

The performance of this decision tree on the entire image set (545 images) was a sensitivity (percentage of correctly classified melanomas) of 69.35% and a specificity (percentage of correctly classified benigns) of 89.97%. The overall classification accuracy for all areas, including structures related to blue-white veil, was 82.94%. This included some areas closely related to blue-white veil such as blue-gray or gray veil. On the subset of images that are known to have blue-white veil areas (44 benigns, 134 melanomas), the sensitivity and specificity were 76.87% and 75.00%, respectively. On the other hand, on the subset of melanomas (133 cases) for which the blue-white veil is the primary feature the sensitivity was 78.20%.

## 6. CONCLUSIONS

In this article, a machine learning approach to the detection of blue-white veil in dermoscopy images was described. The method is comprised of several steps including preprocessing, feature extraction, decision tree induction, rule application, and postprocessing. The detected blue-white areas were characterized using a numerical feature, which in conjunction with an ellipticity measure yielded a sensitivity of 69.35% and a specificity of 89.97% on a set of 545 dermoscopy images. The presented blue-white veil detector takes a fraction of a second for a 768 x 512 image on an Intel Pentium D 2.66GHz. computer.

## 7. ACKNOWLEDGMENTS

This work was supported by grants from NIH (SBIR #2R44 CA-101639-02A2), NSF (#0216500-EIA), Texas Workforce Commission (#3204600182), and James A. Schlipmann Melanoma Cancer Foundation. The permissions to use the images from the CD-ROM Interactive Atlas of Dermoscopy and American Academy of Dermatology DVD on Dermoscopy are gratefully acknowledged.

**FIGURE LEGEND**

**Figure 1.** Melanoma with blue-white veil (a) clinical image and (b) dermoscopy image. The steps of the blue-white veil detection procedure will be demonstrated on image (b).

**Figure 2.** Overview of the approach

**Figure 3.** Preprocessing

**Figure 4.** Preprocessing steps (a) B-spline approximation of the border, (b) binary border mask, (c) 10% (gray) and 20% (white) areas outside the lesion, and (d) manually selected veil (left circle) and non-veil (right circle) regions

**Figure 5.** Pixel classification tree

**Figure 6.** Postprocessing (a) initial veil mask and (b) final veil mask

**Figure 7.** Sample blue-white veil detection results. The veil and non-veil region borders are delineated with thick and thin lines, respectively.

**Figure 8.** Image classification tree

**TABLE LEGEND**

**Table 1.** Description of the color features

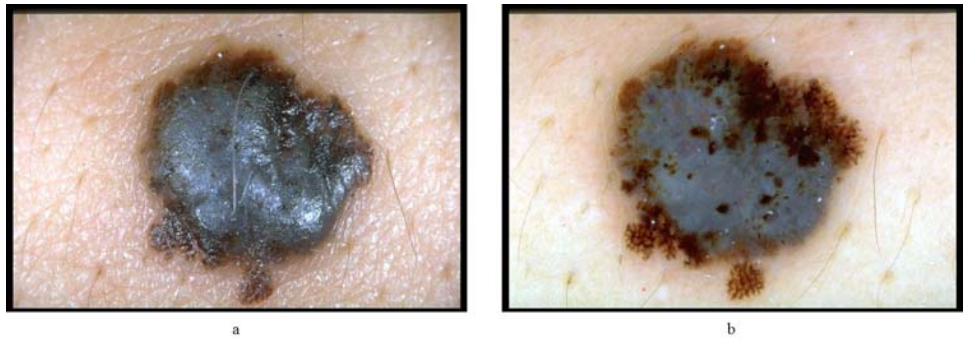

**Figure 1.** Melanoma with blue-white veil (a) clinical image and (b) dermoscopy image. The steps of the blue-white veil detection procedure will be demonstrated on image (b).

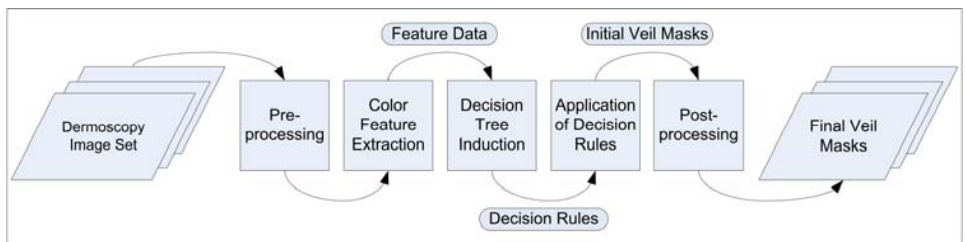

**Figure 2.** Overview of the approach

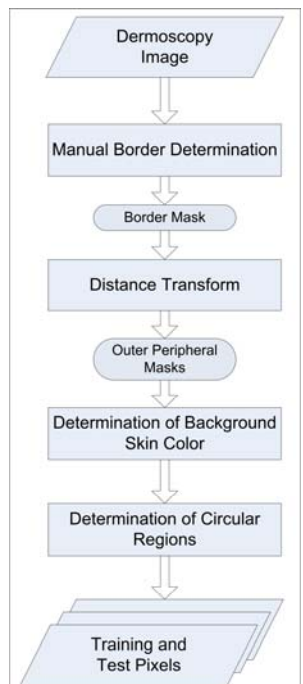

**Figure 3.** Preprocessing

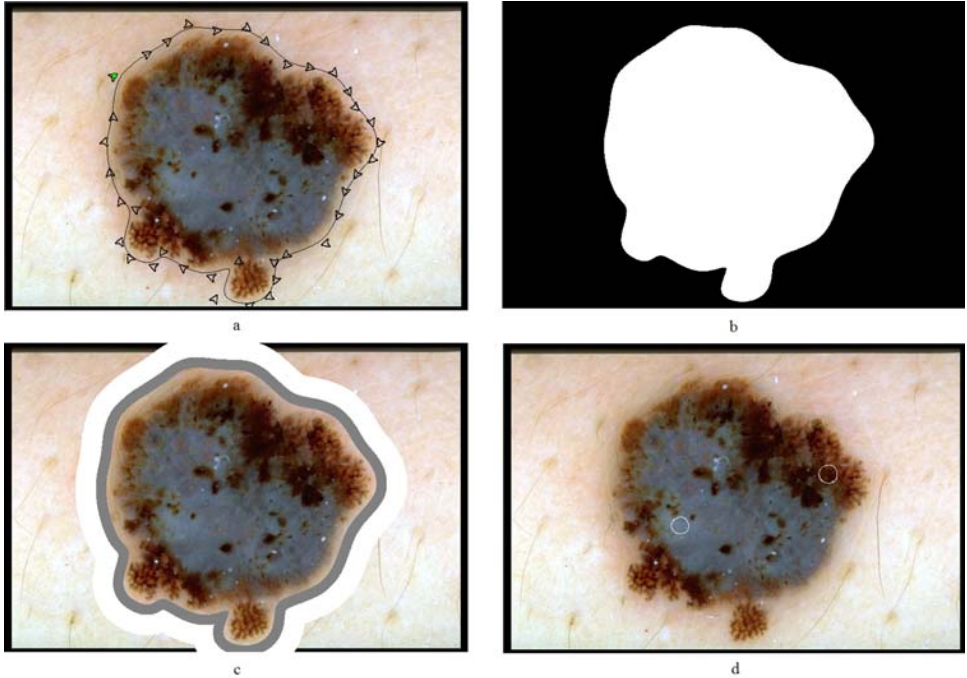

**Figure 4.** Preprocessing steps (a) B-spline approximation of the border, (b) binary border mask, (c) 10% (gray) and 20% (white) areas outside the lesion, and (d) manually selected veil (left circle) and non-veil (right circle) regions

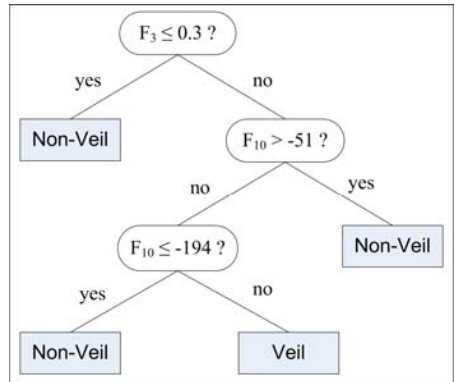

**Figure 5.** Pixel classification tree

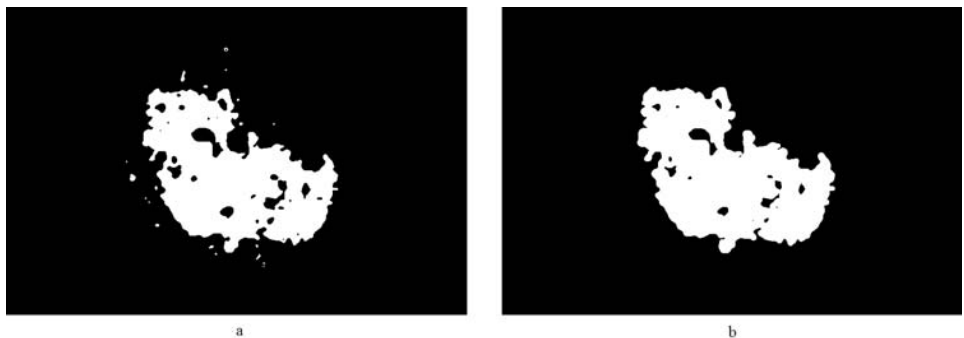

**Figure 6.** Postprocessing (a) initial veil mask and (b) final veil mask

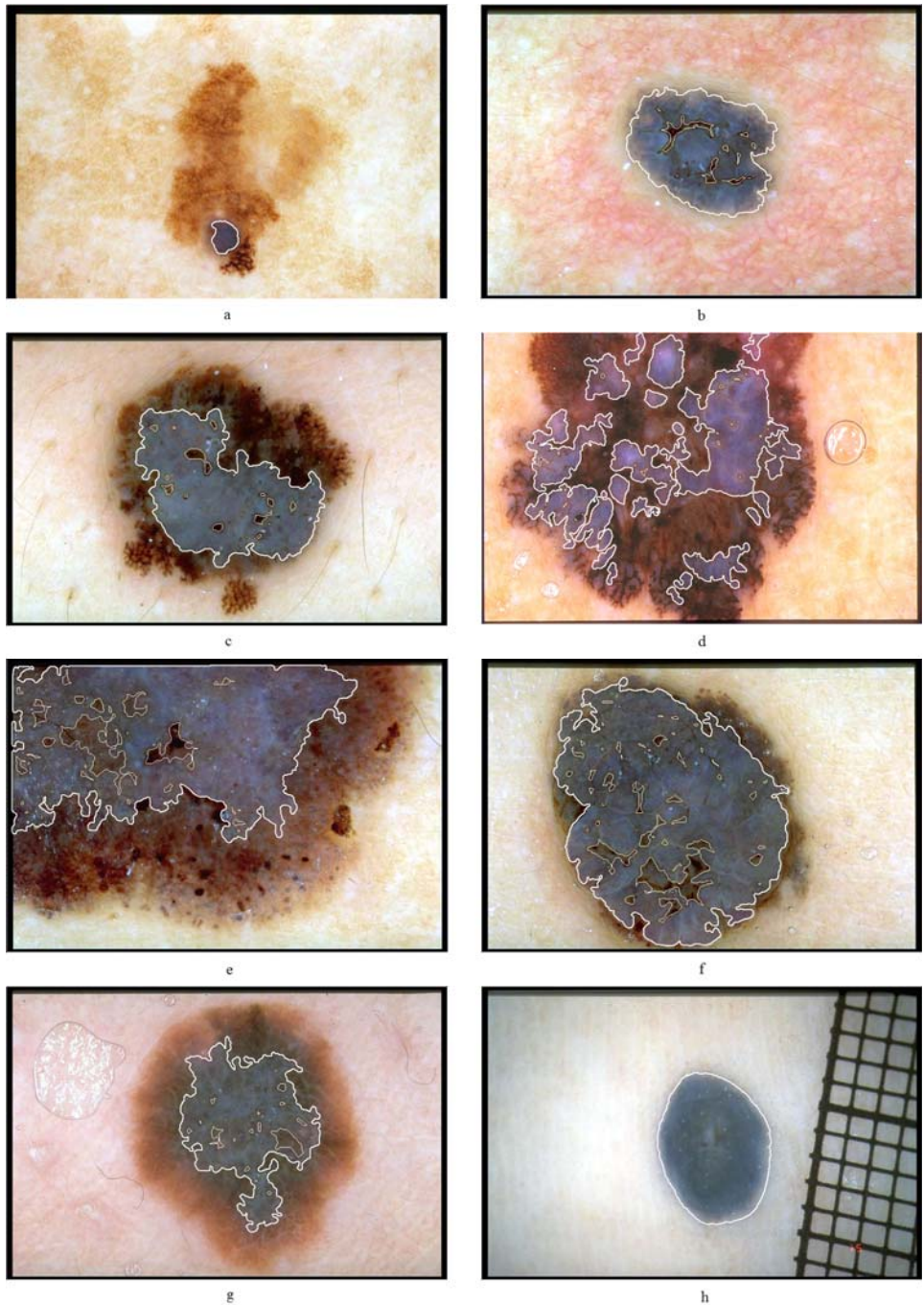

**Figure 7.** Sample blue-white veil detection results. The veil and non-veil region borders are delineated with thick and thin lines, respectively.

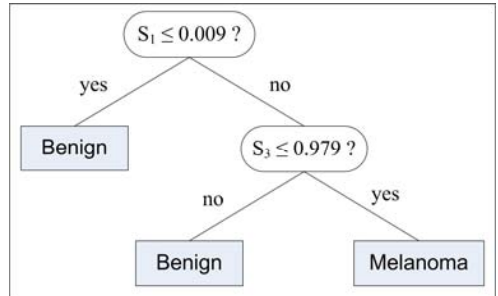

**Figure 8.** Image classification tree

**Table 1.** Description of the color features

| Feature Group | Description |
|---|---|
| $F_1 = \dfrac{R_L}{R_L+G_L+B_L}$ ; $F_2 = \dfrac{G_L}{R_L+G_L+B_L}$ ; $F_3 = \dfrac{B_L}{R_L+G_L+B_L}$ | Chromaticity Coordinates |
| $F_4 = \dfrac{R_L}{R_S}$ ; $F_5 = \dfrac{G_L}{G_S}$ ; $F_6 = \dfrac{B_L}{B_S}$ | Relative $R$, $G$, $B$ Ratio |
| $F_7 = \dfrac{F_4}{F_4+F_5+F_6}$ ; $F_8 = \dfrac{F_5}{F_4+F_5+F_6}$ ; $F_9 = \dfrac{F_6}{F_4+F_5+F_6}$ | Normalized Relative $R$, $G$, $B$ Ratio |
| $F_{10} = R_L - R_S$ ; $F_{11} = G_L - G_S$ ; $F_{12} = B_L - B_S$ | Relative $R$, $G$, $B$ Difference |
| $F_{13} = \dfrac{F_{10}}{F_{10}+F_{11}+F_{12}}$ ; $F_{14} = \dfrac{F_{11}}{F_{10}+F_{11}+F_{12}}$ ; $F_{15} = \dfrac{F_{12}}{F_{10}+F_{11}+F_{12}}$ | Normalized Relative $R$, $G$, $B$ Difference |